\documentclass[11pt]{article}

\usepackage[final]{acl}

\usepackage{times}
\usepackage{latexsym}

\usepackage[T1]{fontenc}

\usepackage[utf8]{inputenc}

\usepackage{microtype}

\usepackage{inconsolata}

\usepackage{graphicx}

\usepackage[table]{xcolor}
\usepackage{subfigure}
\usepackage{multirow}
\usepackage{multicol}
\usepackage{booktabs}
\usepackage{makecell}
\usepackage[most]{tcolorbox}
\usepackage{enumitem}
\usepackage{algorithm}
\usepackage{algorithmic}
\usepackage{amssymb}

\newcommand{\ourM}{\textsc{SciCustom}}

\newcommand{\todo}[1]{{\color{blue}{\textsf{[#1]}}}}
\definecolor{mygray}{gray}{.9}

\title{
SciCustom: A Framework for Custom Evaluation of Scientific Capabilities in Large Language Models
}

\author{
  Yiyang Gu$^{1}$\thanks{Equal contribution with order determined by flipping a coin.}, Junwei Yang$^{1 *}$, Junyu Luo$^{1}$, Ye Yuan$^{1}$, Bin Feng$^{3}$, Yingce Xia$^{2}$,  \\
  \textbf{Shufang Xie$^{2}$, Kaili Liu$^{4}$, Bohan Wu$^{1}$, Qi Shi$^{5}$, Haoran Li$^{5}$, Beier Xiao$^{5}$, Zhiping Xiao$^{6}$\thanks{Corresponding authors.}, }  \\
  \textbf{Xiao Luo$^{7}$, Weizhi Zhang$^{8}$, Philip S. Yu$^{8}$, Zequn Liu$^{2 \dagger}$, Ming Zhang$^{1 \dagger}$} 
  \\
  $^1$ State Key Laboratory for Multimedia Information Processing, School of Computer Science, \\
  PKU-Anker LLM Lab, Peking University
  $^2$ Zhongguancun Academy
  $^3$ IDEA\\
  $^4$ Xidian University
  $^5$ Peking University 
  $^6$ University of Washington \\
  $^7$ University of Wisconsin–Madison
  $^8$ University of Illinois Chicago\\
  \small{
  \texttt{\{yiyanggu,yjwtheonly,mzhang\_cs\}@pku.edu.cn}, \, \texttt{patxiao@uw.edu}, \,
  \texttt{liuzequn@bza.edu.cn}}\vspace{0.5cm}\\
}

\begin{document}
\maketitle

\begin{abstract}
Large language models (LLMs) are increasingly applied to scientific research, yet existing evaluations often fail to reflect the fine-grained capabilities required in practice. Most benchmarks are manually curated or domain-generic, limiting scalability and alignment with real scientific use cases.
In this paper, we propose a new framework named \ourM{} to address the problem. It enables the custom construction of benchmarks from large-scale scientific data to evaluate application-specific scientific capabilities in LLMs. \ourM{} first organizes scientific knowledge into ontology-grounded knowledge units with controlled granularity and trains a tagger to map large-scale data instances into this knowledge space. Given a custom requirement, relevant knowledge units are identified via voting-based multi-model consensus. These units enable relevance-aware benchmark retrieval via binary search, followed by proxy subset selection and data-grounded benchmark generation for efficient evaluation.
Experiments in chemistry and healthcare demonstrate that \ourM{} reveals fine-grained differences in LLM scientific capabilities that standard benchmarks overlook, while requiring neither expert annotation nor synthetic question generation. 
This work provides a scalable and application-aware foundation for benchmarking scientific capabilities in LLMs. The source code is available at \url{https://github.com/yjwtheonly/SciCustom}.
\end{abstract}

\section{Introduction}

\begin{figure}[t]
\centering	\includegraphics[width=.95\linewidth]{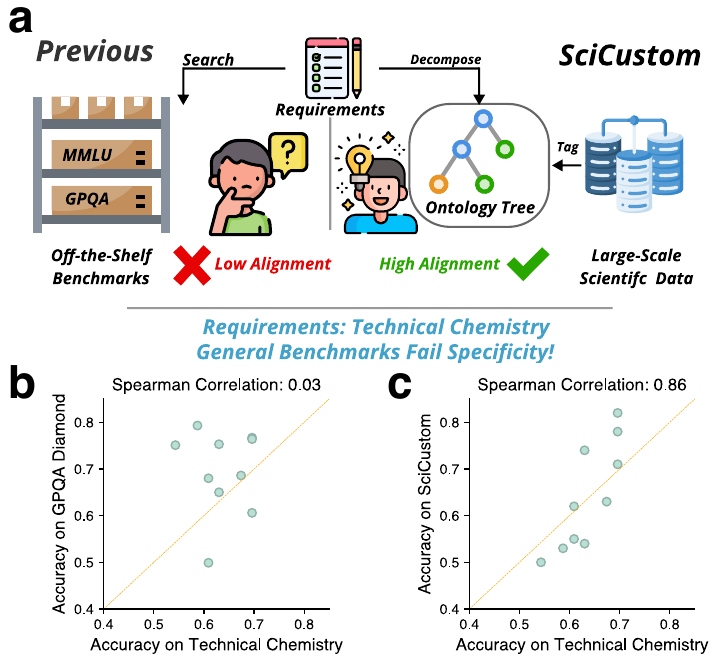}
	\caption {
    Illustrations of \ourM{}. (\textbf{a}) Comparison between traditional off-the-shelf benchmarking and our ontology-driven framework.
    (\textbf{b, c}) Evaluation of 10 LLMs on different benchmarks, where each dot represents a model. Targeting specific capabilities in Technical Chemistry, (\textbf{b}) the general scientific benchmark (GPQA Diamond) aligns poorly with expert ground truth, whereas (\textbf{c}) the benchmark constructed by \ourM{} demonstrates strong alignment.
    }
\label{fig::overview}
\vspace{-4mm}
\end{figure}

With the rapid advancement of Large Language Models (LLMs), their applications have significantly broadened across the scientific landscape~\cite{Chang2024,Bommasani2021FoundationModels,Birhane2023ScienceLLMs,luo2025large,yuan2024measuring,liu-etal-2023-molxpt,xia2025naturelanguagemodeldeciphering}. For example, an LLM-based system successfully proposed a novel mechanism of gene transfer in bacteria, mirroring conclusions that took years of experimental validation~\cite{Gottweis2025,Penades2025}. However, as scientists increasingly seek to leverage LLMs for specific scientific applications, a critical challenge emerges: the performance of a given LLM within a particular scientific context remains largely uncertain and difficult to evaluate~\cite{Cai2025,Singhal2025MedPaLM2,Miret2025MatSciLLM,bedi2025testing}.

While numerous benchmarks exist to evaluate various aspects of model capability, they frequently fail to reflect the requirements of specialized usage~\cite{Anjum2025,Liang2023HELM,Bandel2024Unitxt}. Empirical observations indicate that performance on widely-used benchmarks often diverges from that in specific scientific tasks (\textbf{Figure \ref{fig::overview}b}), necessitating the customization of benchmarks tailored to particular scientific applications~\cite{Singhal2025MedPaLM2,Miret2025MatSciLLM}.
Given the vast and ever-expanding range of scientific applications, manual curation of benchmarks for each emerging use case is impractical~\cite{Huang2021TDC,Wu2018MoleculeNet}.

A plausible solution is to automate the customized benchmark construction process~\cite{Farchi2024,autobencher,chou2025autocodebench,wang2025omnieval}. However, developing such an automated framework presents two challenges. First, scientific applications are inherently complex and highly interdisciplinary. A single application often requires knowledge from multiple subfields (e.g., drug discovery intertwines organic chemistry, molecular biology, and pharmacology~\cite{Lu2025}), while distinct applications frequently share underlying knowledge (e.g., both drug discovery and clinical decision-making require the knowledge of pharmacology~\cite{song2025llm,bi2025machine,Ong2025}). Consequently, constructing benchmarks from scratch for each specific scenario results in massive redundant labor and limits the framework's scalability~\cite{Huang2021TDC,li2025dmpkbench}. Second, scientific evaluation requires grounded validity. High-quality evaluation data are frequently derived from costly wet-lab experiments or computational simulations~\cite{chen2024validation,ramos2025review}, rendering simple general-purpose data synthesis methods infeasible~\cite{Parker2020,wei2024magicoder,chou2025autocodebench,xu2025kodcode}.

To this end, we introduce \ourM{}, a framework for custom evaluation of scientific capabilities in LLMs (\textbf{Figure \ref{fig::overview}a,c}). The key idea is that a complex scientific application can be approximated by a composition of fine-grained knowledge units. To construct these units, we select 642 representative concepts from a comprehensive scientific ontology to serve as unified knowledge units. Leveraging the rich hierarchical information within the ontology, which links specific scientific concepts to knowledge units, we train a tagger to automatically align real-world scientific corpora with the corresponding knowledge units. This offline process performs a once-for-all population of each knowledge unit with grounded scientific data. When a user presents an evaluation requirement, \ourM{} identifies the relevant knowledge units and orchestrates a hierarchical retrieval process to construct a tailored benchmark. The combinability of these pre-constructed knowledge units enables dynamic adaptation to new requirements via efficient reuse, ensuring scalability for diverse downstream applications.

Experimental results in chemistry and healthcare domains demonstrate that our automatically constructed benchmarks closely align with expert-curated benchmarks. Furthermore, for pericyclic reaction, a novel scientific application lacking prior benchmarks, our framework successfully generates a high-quality benchmark, providing a scalable and reliable solution for assessing LLMs in the ever-expanding landscape of scientific research.

Our contributions can be summarized as follows:
\begin{itemize}
    \item We propose \ourM{}, a framework that models scientific applications as compositions of reusable knowledge units to automate benchmark construction, enabling efficient adaptation to diverse downstream tasks.
    
    \item We develop a tagger that leverages ontology structures to map large-scale, real-world scientific corpora into knowledge units, ensuring that evaluations are grounded in verifiable data.

    \item Experiments in chemistry and healthcare confirm \ourM{}'s strong alignment with expert-curated benchmarks and its capability to evaluate novel scientific contexts where no prior benchmarks exist.
    
\end{itemize}

\section{Methodology}

\begin{figure*}[t]
\centering	\includegraphics[width=.98\linewidth]{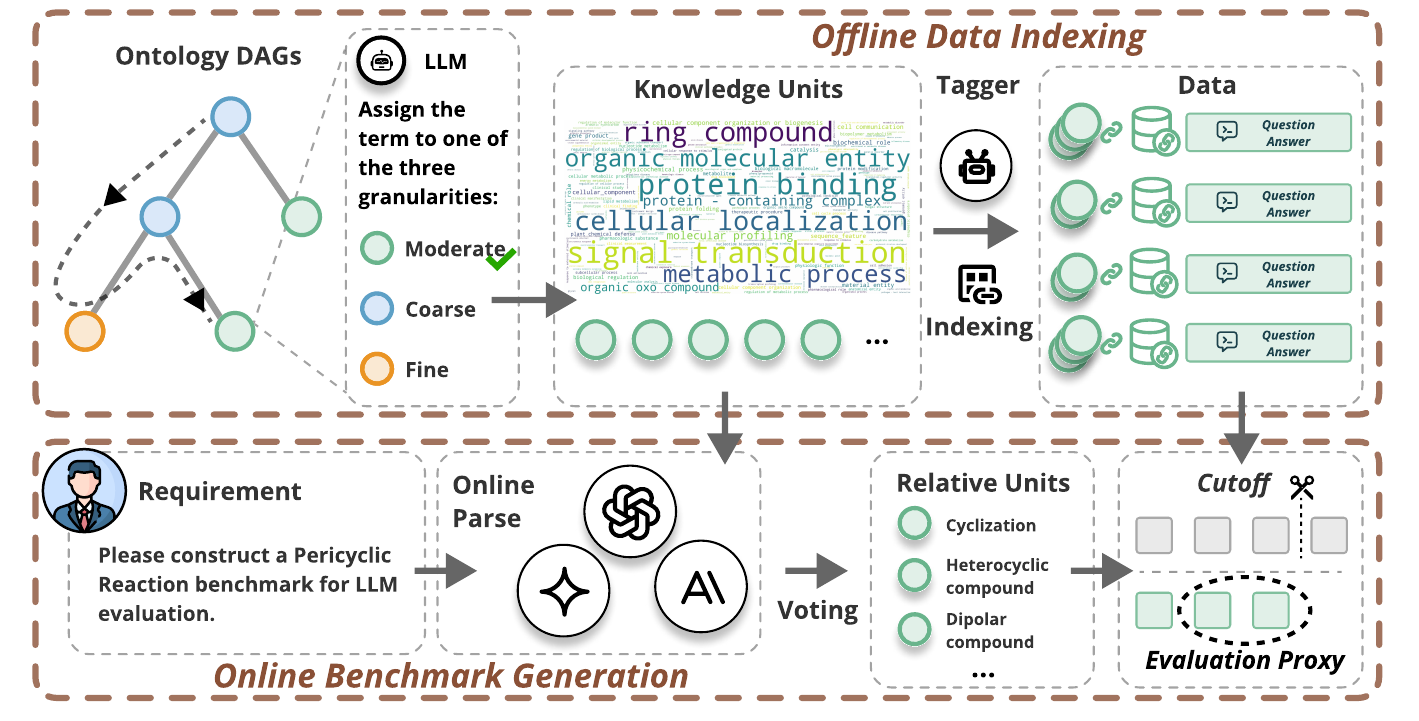}
    \vspace{-4mm}
	\caption {Framework of \ourM{}. It consists of an offline phase where scientific data is indexed into ontology-grounded knowledge units via a trained tagger, and an online phase where user requirements are parsed by multi-model voting to identify relevant tags. These tags guide the binary search-based selection and proxy selection of data. The problem set of the benchmark is generated based on these data.}
    \vspace{-3mm}
\label{fig::framework}
\end{figure*}

\subsection{Overview}
We study the problem of customized scientific capability evaluation for LLMs. Given an evaluation requirement $r$ and a large corpus of scientific data $\mathcal{D}$, the goal is to construct a benchmark $\mathcal{B}_r$ that reflects the capabilities required by $r$ without expert annotation.

\ourM{} is an automated framework for this problem, operating through offline data indexing and online data-grounded generation phases. In the offline phase, \ourM{} represents large-scale scientific data within shared knowledge units using a tagger. This phase organizes scattered scientific content into reusable units. 
In the online phase, \ourM{} first employs a voting-based multi-model consensus to identify knowledge units relevant to the user requirement $r$. Guided by these units, the framework performs a hierarchical data selection process: retrieving candidate data, filtering via binary search, and extracting a representative proxy subset to generate final questions for efficient evaluation. 
An overview of \ourM{} is depicted
in \textbf{Figure~\ref{fig::framework}}.

\subsection{Ontology-Grounded Knowledge Units}
\label{sec:ontology_anchors}

To enable fine-grained yet reusable modeling of scientific knowledge, \ourM{} is built upon a structured scientific ontology that serves as a unified semantic backbone for both knowledge definition and data organization. 

Following \citet{liu2021graphine}, we integrate multiple authoritative repositories to construct a large-scale scientific ontology, organized into a collection of directed acyclic graphs (DAGs) spanning $227$ scientific subdisciplines. Within each DAG, concepts are interconnected via \emph{is-a} relations, where descendant nodes represent finer-grained specializations of their parent concepts (e.g., \textit{Organic Chemistry} is a descendant node of \textit{Chemistry}.) 

We select concepts from this ontology as knowledge units. The knowledge units should be neither overly coarse nor excessively specific. Overly coarse concepts lack the resolution to distinguish fine-grained model capabilities, while excessively specific ones hinder the effective reuse of knowledge units for broader applications. We empirically select concepts at a granularity comparable to ``textbook chapter titles''.
We therefore capture these knowledge units via a depth-first traversal over each ontology DAG $\mathcal{G}_i$. 
At each node $v$, an LLM evaluates whether the ontology term aligns with the granularity criteria based on the term name, illustrative examples, and its prior knowledge of scientific taxonomy.  

\begin{algorithm}[tb]
    \caption{Ontology-Guided Knowledge Unit Selection}
    \label{alg:anchor_selection}
    \textbf{Input}: Ontology $\{\mathcal{G}_i\}$, \textbf{Output}: Units $\mathcal{T}$
    \begin{algorithmic}[1]
    \STATE $\mathcal{T} \leftarrow \emptyset$
    \FOR{\textbf{each} node $v$ visited by DFS in $\{\mathcal{G}_i\}$}
        \IF{$|\mathrm{Desc}(v)| < 10$} 
            \STATE Backtrack 
        \ENDIF
        \STATE $label \leftarrow$ LLM classifies $v$ as \textit{coarse}, \textit{moderate}, or \textit{fine}
        \IF{$label$ is \textit{coarse}}
            \STATE Continue traversal (recurse into children)
        \ELSIF{$label$ is \textit{moderate}}
            \STATE $\mathcal{T} \leftarrow \mathcal{T} \cup \{v\}$
        \ELSE 
            \STATE Backtrack \COMMENT{Prune branch if \textit{fine}}
        \ENDIF
    \ENDFOR
    \STATE \textbf{return} $\mathcal{T}$
    \end{algorithmic}
\end{algorithm}

\begin{figure}[t]
\centering	\includegraphics[width=.98\columnwidth]{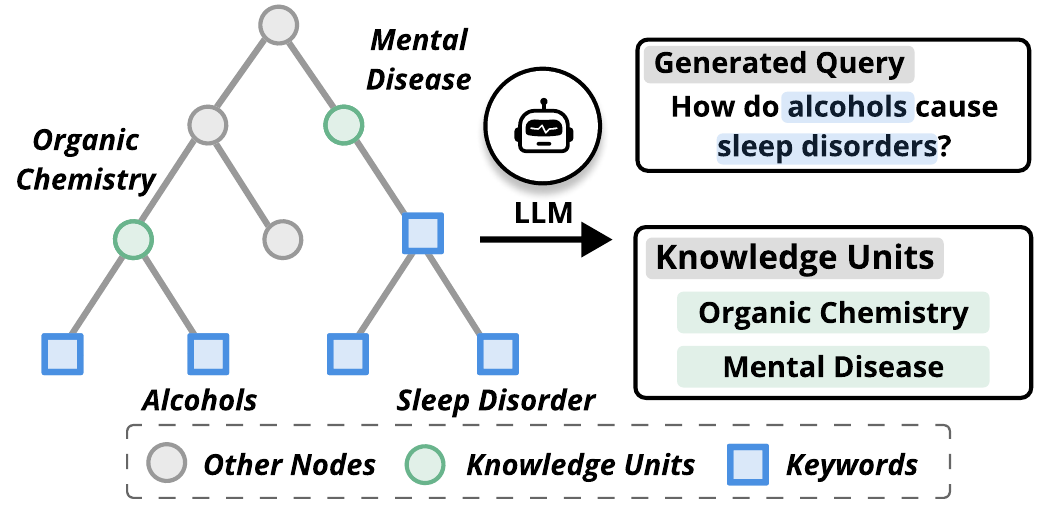}
	\caption {Illustration of the synthetic data construction pipeline for tagger training. We sample knowledge units (green circles) and extract descendant keywords (blue squares) from the ontology. An LLM then generates a natural language query based on these keywords, creating a labeled training instance.}
\label{fig::tagger}
\vspace{-4mm}
\end{figure} 

The overall procedure is summarized in Algorithm~\ref{alg:anchor_selection}, and the detailed prompting strategy is provided in Appendix~\ref{prompt::subfield-granularity}.

Through this procedure, we obtain $641$ scientific knowledge units. We additionally introduce a dedicated \textsc{Non-Scientific} unit, resulting in a total of $642$ knowledge units. These units define the ontology-grounded scientific knowledge space that underpins all subsequent benchmark construction in \ourM{}.

\subsection{Mapping Data to Knowledge Units}
\label{sec:ontology_tagger}
To populate the identified knowledge units with grounded scientific data, we collect a large-scale corpus containing high-quality question-answering (QA) pairs across diverse scientific domains. The corpus comprises a total of $N$ data instances, each defined as $d = \langle q, a \rangle$, where $q$ denotes a natural language question and $a$ the corresponding reference answer. These instances are originally scattered, necessitating the development of a specialized scientific tagger to organize them into knowledge units.

To enable efficient large-scale annotation, we train a small language model as the tagger. The model maps each query $q$ to a subset of knowledge units, $\mathcal{T}_{q} \subseteq \mathcal{T}$. 
This design implements a once-for-all mapping between data and knowledge units during the construction phase, allowing the same annotated corpus to be reused across diverse evaluation requirements.

Supervised data for training the tagger is constructed using a two-stage strategy. 
In the first stage, we generate synthetic scientific queries via controlled compositional sampling over the knowledge units $\mathcal{T}$. Specifically, we randomly sample between $1$ and $5$ knowledge units to form a target set $\mathcal{T}_{q} \subset \mathcal{T}$. We fully leverage the rich hierarchical information of the ontology: for each selected unit $v \in \mathcal{T}_{q}$, we aggregate all its descendant nodes to form a representative keyword set $\mathcal{K}_v$ (e.g., \textit{Ventricular Septum} is a keyword for the unit \textit{Anatomical Entities}). We further sample keywords from each $\mathcal{K}_v$ and combine them to form a composite prompt for an LLM to generate a synthetic query $q$ (\textbf{Figure \ref{fig::tagger}}). Each generated query is then paired with its corresponding set of source knowledge units, yielding a labeled training instance of the form $\langle q, \mathcal{T}_{q} \rangle$. Using this procedure, we construct synthetic scientific training instances covering complex compositional knowledge patterns. In the second stage, we complement the synthetic data with real-world scientific queries sampled from existing instruction-tuning datasets. These queries are annotated with their corresponding knowledge units with an LLM. In addition, we include non-scientific queries to model boundary cases between scientific and non-scientific content.

By introducing a small and efficient tagger for annotating scientific queries, \ourM{} achieves scalable and interpretable alignment between natural language queries and scientific knowledge, forming a key foundation for customized scientific evaluation.

\subsection{Voting-Based Knowledge Unit Selection}
\label{sec:voting}

With the large-scale scientific corpus successfully grounded to ontology-defined knowledge units via the tagger, we possess a structured knowledge base ready for dynamic retrieval. To construct a benchmark tailored to a specific user evaluation requirement $r$, the critical first step is to accurately identify the subset of knowledge units $\mathcal{T}_r \subset \mathcal{T}$ that underpins the requirement.

To identify the target knowledge units $\mathcal{T}_r$, we employ a multi-model consensus mechanism where a set of heterogeneous LLMs independently rank candidate units based on their relevance to $r$. We aggregate these rankings by calculating the average rank position for each unit across all models, and select the top-$K_1$ units with the lowest consensus ranks (indicating highest relevance) to form the target set $\mathcal{T}_r$.

\subsection{Hierarchical Benchmark Generation}
\label{sec:benchmark generation}
\paragraph{Binary Search-Based Data Selection.}
With the requirement-relevant knowledge unit set $\mathcal{T}_r$ established, we aim to retrieve a dataset $\mathcal{D}_r$ from the massive corpus $\mathcal{D}$ that best matches $r$. For any data sample $\langle q, a \rangle \in \mathcal{D}$, let $\mathcal{T}_{r,q} = \mathcal{T}_r \cap \mathcal{T}_q$ represent the intersection between the data's inherent knowledge units and the query knowledge units.

Intuitively, we construct a candidate set $\mathcal{D}_r' = \{ \langle q, a \rangle \in \mathcal{D} \mid |\mathcal{T}_{r,q}| \ge 1 \}$. However, a simple intersection filter is insufficient to guarantee high semantic alignment. While employing LLMs to verify the relevance of each candidate against $r$ would ensure quality, applying this to the entire set is computationally prohibitive. To optimize API usage without compromising retrieval quality, we adopt a binary search-based strategy that relies on a pre-sorted candidate list. Specifically, we define a ordering where data sample $d_i = \langle q_i, a_i \rangle$ is ranked higher than $d_j = \langle q_j, a_j \rangle$ if:
\begin{enumerate}
    \item $|\mathcal{T}_{r, q_i}| > |\mathcal{T}_{r, q_j}|$; or
    \item $|\mathcal{T}_{r, q_i}| = |\mathcal{T}_{r, q_j}|$ and $\bar{\mathcal{R}}_r(d_i) < \bar{\mathcal{R}}_r(d_j)$,
\end{enumerate}
where $\mathcal{T}_{r,q} = \mathcal{T}_r \cap \mathcal{T}_q$ denotes the matching knowledge units, and $\bar{\mathcal{R}}_r(d)$ represents average rank of the knowledge units in $\mathcal{T}_{r,q}$ acquired in Sec. \ref{sec:voting}.

\begin{algorithm}[t]
\caption{Binary Search for Relevant Benchmark Construction}
\label{alg:binary_search}
\textbf{Input}: Sorted list $L_{\mathcal{D}'_r}$, Requirement $r$, Models $\mathcal{M}$ \quad \textbf{Output}: Data $\mathcal{D}_r$

\begin{algorithmic}[1]
\STATE $low \leftarrow 0, \ high \leftarrow |L_{\mathcal{D}'_r}| - 1, \ cutoff \leftarrow 0$
\WHILE{$low \leq high$}
    \STATE $mid \leftarrow low + \lfloor (high - low) / 2 \rfloor$; \quad $d_{mid} \leftarrow L_{\mathcal{D}'_r}[mid]$
    \STATE $vote \leftarrow \sum_{M_i \in \mathcal{M}} \mathbb{I}(M_i \text{ judges } d_{mid} \text{ relevant to } r)$
    \IF{$vote > |\mathcal{M}| / 2$}
        \STATE $cutoff \leftarrow mid$; \quad $low \leftarrow mid + 1$
    \ELSE
        \STATE $high \leftarrow mid - 1$
    \ENDIF
\ENDWHILE
\RETURN $\mathcal{D}_r = L_{\mathcal{D}'_r}[0 : cutoff]$
\end{algorithmic}
\end{algorithm}

We assume that within the prioritized candidate list, the relevance to the requirement $r$ follows a generally monotonic decreasing trend. 
Under this assumption, we apply the binary search procedure to efficiently identify the cutoff point—defined as the last position where a majority of the LLM ensemble still judges the sample as relevant to $r$.
This approach enables us to determine the maximal relevant data subset without linearly verifying every instance, significantly reducing the number of required oracle judgments to $O(\log(|\mathcal{D}'_r|))$.
The detailed procedure is outlined in Algorithm \ref{alg:binary_search}.

\paragraph{Efficient Proxy Subset Selection.}
The constructed dataset $\mathcal{D}_r$ may still be excessively large (e.g., $|\mathcal{D}_r| > 200,000$ in our experiments).
To enable efficient evaluation, we propose a selection strategy to extract a proxy subset $\mathcal{P}_r \subseteq \mathcal{D}_r$ such that $|\mathcal{P}_r| \leq K_2$, while maintaining the evaluative power of the full set.

For each data sample $\langle q, a \rangle \in \mathcal{D}_r$, we compute two commonly used intrinsic metrics~\cite{saranathan2025sublime}: hardness score characterizing the difficulty and quality score indicating the linguistic quality. 
Our objective is to select $\mathcal{P}_r$ such that the distributions of these scores closely approximate those of $\mathcal{D}_r$. We formulate this as minimizing the cumulative Wasserstein distance between the distributions of the subset and the full set. 
To solve the sampling of $\mathcal{P}_r$ efficiently, we adopt a cluster-based strategy following SubLIME~\cite{saranathan2025sublime}.
Intuitively, we encode all pairs $d = \langle q, a \rangle$ into an embedding space using RoBERTa~\cite{liu2019roberta}.
We then perform clustering in embedding space and sample representatives from each cluster. Detailed information is in Appendix \ref{app:proxy}.

\paragraph{Data-Grounded Benchmark Generation.}
To facilitate standardized and deterministic evaluation, we further transform the raw query-answer pairs in the selected proxy subset $\mathcal{P}_r$ into multiple-choice questions (MCQs). We utilize LLM to generate plausible distractors for each instance. The final output is a structured, efficient benchmark $\mathcal{B}_r$ composed of high-quality MCQs, ready for immediate automated model assessment.
\section{Experiments}
We established a comprehensive experimental framework to evaluate the effectiveness of \ourM{}. First, we benchmarked our approach against groundtruth expert-curated datasets across 11 specific evaluation requirements. We focused on measuring the alignment of model rankings produced by \ourM{} versus those from human experts. Furthermore, to verify the framework's utility in real-world exploration, we applied \ourM{} to Pericyclic
Reaction, a novel evaluation requirement without existing benchmarks, and qualitatively analyzed the validity of the constructed benchmark.
\subsection{Experimental Setup}

\begin{table*}[t]
\centering
\caption{Ranking consistency analysis of 10 LLMs on 6 chemistry-specific benchmarks. The table compares \ourM{} against baseline benchmarks in preserving model rankings, quantified by Spearman and Kendall correlation coefficients.
Higher scores indicate better consistency. Best results are in \textbf{bold}.}
\vspace{-1mm}
\footnotesize
\resizebox{\linewidth}{!}{
\begin{tabular}{c|c|cccccc}
\toprule
 Metrics & Benchmarks    &   Analytical chemistry &   Inorganic chemistry &   Material science &  Organic chemistry &  Physical chemistry &    Technical chemistry \\
 \midrule
 \multirow{8}{*}{
 \rotatebox{90}{
    \makecell{Spearman\\correlation}
  }
 }
 & IfBench  & -0.32   & -0.34 & -0.43 &  0.12 &  -0.61 &  -0.54 \\
 & SimpleQA & -0.67   & -0.69 & -0.78 & -0.48 &  -0.42 &  -0.31 \\
  & GPQA     &  0.61  &  0.52 &  0.21 &  0.72 &  0.21  &   0.03 \\ 
  & MMLU     &  0.21  &  0.27 & -0.61 &  0.21 &  0.52  &   0.31 \\
  & MedQA    &  0.31  &  0.42 &  0.01 & -0.21 &  0.63  &   0.72 \\
  & GPT-5    & -0.11  &  0.05 & -0.04 & 0.38  &  0.24  &   -0.07\\ 
  & Embedding& -0.34 &  -0.59 & -0.39 &  0.11 &  -0.73 &  -0.41\\
 & \cellcolor{mygray}{{\ourM}} 
 & \cellcolor{mygray}{\textbf{0.86}}
 & \cellcolor{mygray}{\textbf{0.67}}
 & \cellcolor{mygray}{\textbf{0.42}}
 & \cellcolor{mygray}{\textbf{0.89}}
 & \cellcolor{mygray}{\textbf{0.74}}
 & \cellcolor{mygray}{\textbf{0.86}}\\
  \midrule
 \multirow{8}{*}{
 \rotatebox{90}{
    \makecell{Kendall\\correlation}
  }
 }
 & IfBench    & -0.14 & -0.19 & -0.29 & 0.10  & -0.43  & -0.43  \\
 & SimpleQA   & -0.52 & -0.49 & -0.56 & -0.39 & -0.24  & -0.23  \\
  & GPQA      & 0.39  & 0.39  & 0.10  & 0.42 &  0.14 & -0.05  \\
  & MMLU      & 0.14  & 0.18  & -0.49 & 0.10 &  0.43 & 0.24  \\
  & MedQA     & 0.24  & 0.40  & -0.09 & -0.14 &  0.43 & 0.48  \\
  & GPT-5     & -0.15 & 0.00  & -0.06 & 0.35  & 0.16  & 0.00\\
  & Embedding & -0.31 & -0.52 & -0.31 & 0.09  & -0.62 & -0.32\\
 & \cellcolor{mygray}{{\ourM}} 
 & \cellcolor{mygray}{\textbf{0.62}}
 & \cellcolor{mygray}{\textbf{0.48}}
 & \cellcolor{mygray}{\textbf{0.35}}
 & \cellcolor{mygray}{\textbf{0.68}}
 & \cellcolor{mygray}{\textbf{0.57}}
 & \cellcolor{mygray}{\textbf{0.65}}\\
\bottomrule
\end{tabular}
}
\label{table::main_results_chemistry}
\end{table*}

\begin{table}[t] 
\centering
\caption{Ranking consistency analysis of 10 LLMs on 5 healthcare-specific benchmarks. Higher scores indicate better consistency. Best results are in \textbf{bold}.}
\vspace{-1mm}
\footnotesize
\setlength{\tabcolsep}{1.5pt} 
\resizebox{\columnwidth}{!}{ 
\begin{tabular}{c|c|ccccc}
\toprule
  Metrics & Benchmarks    &   Virology &   Human aging &  Medical genetics &  Anatomy &  Nutrition \\
 \midrule
  \multirow{8}{*}{
 \rotatebox{90}{
    \makecell{Spearman\\correlation}
  }
 }
 & IfBench &  0.04   &  -0.32 &  -0.64 &  -0.26 &  -0.11 \\
 & SimpleQA &  0.35  & -0.50 &   0.00 &  -0.37 &  0.11  \\
 & GPQA     &  -0.11 &-0.10 &   -0.09 &  0.48 &  0.18  \\
 & MMLU &  0.35 &   0.21 &   0.04 &  -0.15 &  0.56     \\
 & MedQA &  0.44 &   \textbf{0.62} &  0.35  &  -0.19 &  0.45 \\
 & GPT-5 &  0.25 &  0.20 & 0.09 & 0.11 & 0.52 \\
  & Embedding & 0.18 & 0.21 & -0.21& -0.32 & 0.27 \\
 & \cellcolor{mygray}{{\ourM}} 
 & \cellcolor{mygray}{\textbf{0.55}}
 & \cellcolor{mygray}{0.49}
 & \cellcolor{mygray}{\textbf{0.42}}
 & \cellcolor{mygray}{\textbf{0.62}}
 & \cellcolor{mygray}{\textbf{0.78}}\\
\midrule 
  \multirow{8}{*}{
 \rotatebox{90}{
    \makecell{Kendall\\correlation}
  }
 }
 & IfBench &  -0.05   &  -0.14 &  -0.43 &  -0.21 &  -0.00 \\
 & SimpleQA &  0.31  & -0.33 &   0.00 &  -0.21 &  0.10  \\
 & GPQA     &  -0.05 & -0.10 &   -0.11 &  0.32 &  0.10  \\
 & MMLU     &  0.31  &  0.14 &    0.00 &  0.00 &  0.51  \\
 & MedQA    &  0.35  &  \textbf{0.42} &    0.23 &  -0.10 & 0.31  \\
  & GPT-5    & 0.20 & 0.18 & 0.06 & 0.06 & 0.43 \\
  & Embedding &  0.11 &  0.10 &  -0.14 &  -0.16 & 0.18 \\
 & \cellcolor{mygray}{{\ourM}} 
 & \cellcolor{mygray}{\textbf{0.45}}
 & \cellcolor{mygray}{0.37}
 & \cellcolor{mygray}{\textbf{0.33}}
 & \cellcolor{mygray}{\textbf{0.51}}
 & \cellcolor{mygray}{\textbf{0.64}}\\
\bottomrule
\end{tabular}
}
\vspace{-2mm}
\label{table::main_results_biology}
\end{table}
\paragraph{Ground-Truth Benchmarks.}
We evaluated \ourM{} on 6 tasks in chemistry, including analytical chemistry, inorganic chemistry, material science, organic chemistry, physical chemistry and technical chemistry, and 5 tasks in healthcare, including virology, human aging, medical genetics, anatomy and nutrition. The groundtruth benchmarks for chemistry tasks are from \textit{ChemBench} \cite{chembench}, while the healthcare tasks are sourced from the health subset of \textit{MMLU-Pro} \cite{mmlu-pro,MMLU}.

\paragraph{Baselines.} 
We compared \ourM{}-curated benchmarks against a comprehensive set of baselines spanning general-purpose benchmarks, scientific benchmarks and alternative benchmark construction methods. To assess the correlation between general-purpose benchmarks and target tasks, we utilized \textit{IfBench} \cite{ifbench} and \textit{SimpleQA} \cite{simpleqa}. \textit{GPQA-Diamond} \cite{gpqa} was included to compare general scientific benchmarks. For domain-specific comparisons, we selected the chemistry-related subset of \textit{MMLU-Pro} \cite{mmlu-pro} for chemistry tasks and \textit{MedQA} \cite{medqa} for healthcare tasks. Furthermore, to isolate the contributions of our grounded data and ontology-driven design, we introduced two alternative construction baselines: \textit{GPT-5}, a fully synthetic benchmark where the model generates multiple-choice questions directly without access to grounded data, and \textit{Embedding}, an embedding-based baseline that selects data instances from the corpus $\mathcal{D}$ solely via k-nearest neighbor (k-NN) retrieval to the requirement, bypassing the ontology-grounded knowledge units.

\paragraph{Evaluation Protocol.} 
We first evaluate 10 representative LLMs on the ground-truth benchmark to obtain the reference ranking. We then evaluate the same 10 LLMs using benchmarks constructed by each baseline method (including ours) and compute the ranking induced by each alternative benchmark. Finally, we measure the consistency between the baseline-induced ranking and the ground-truth ranking using Spearman and Kendall correlation coefficients.
Further implementation details are provided in Appendix~\ref{sec::supp_implementation}.

\begin{figure*}[t]
\centering
\includegraphics[width=.95\linewidth]{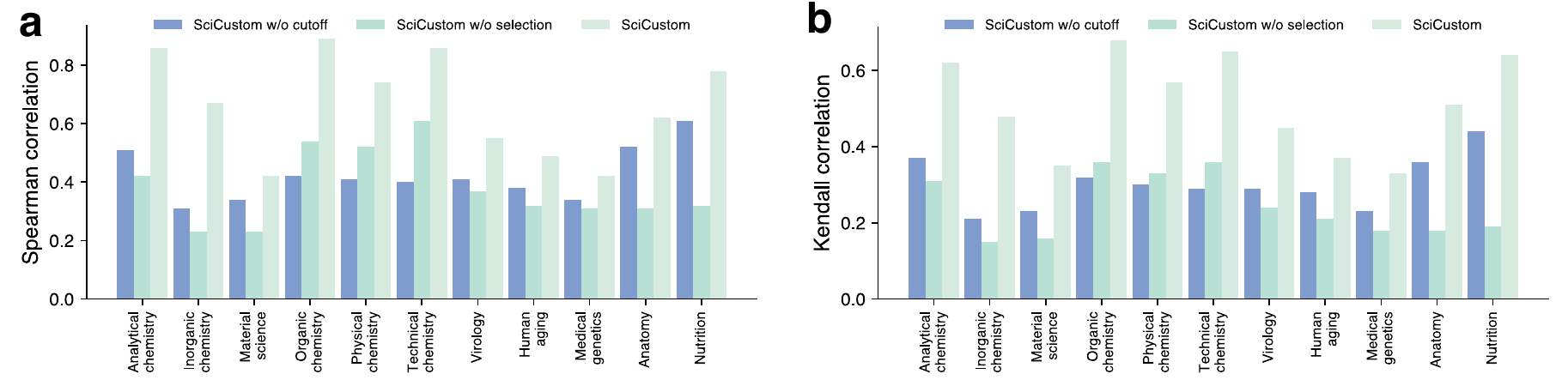}
\vspace{-4mm}
\caption{
Bar plot showing the effectiveness of relevance cutoff and subset selection strategies.
\vspace{-5mm}
}
\label{fig::ablation_slection}
\end{figure*}

\subsection{Alignment Between \ourM{} and Expert-Curated Benchmarks}

We report the results of \ourM{} against various baselines in \textbf{Table \ref{table::main_results_chemistry}} and \textbf{Table \ref{table::main_results_biology}}.
\ourM{} exhibits a robust correlation with expert assessments, achieving the highest Spearman correlation in 10 out of 11 tasks. 
This strong alignment indicates that the benchmarks generated by \ourM{} effectively capture the capabilities required for specific applications, serving as a reliable proxy for expensive expert evaluation. Notably, \ourM{} selects the same top-1 model as the ground-truth benchmark on 8 out of 11 tasks, highlighting its high practical utility in assisting users to identify the optimal model for their specific needs. 

In contrast, widely adopted benchmarks fails to show consistent alignment with expert rankings. This underscores that neither general instruction following skills nor broad scientific reasoning capabilities are reliable predictors of proficiency for specialized scientific requirements.
Furthermore, established domain-specific benchmarks also struggle to predict sub-domain-level performance due to insufficient granularity. These findings further necessitate the \ourM{} framework, which dynamically constructs benchmarks at the precise granularity of the target requirement. 

Comparing \ourM{} with alternative construction baselines reveals the critical role of our methodology design. First, our framework significantly outperforms the fully synthetic \textit{GPT-5} baseline, underscoring the necessity of grounded scientific data. Second, \ourM{} also surpasses the \textit{Embedding} baseline, which retrieves grounded data via simple semantic similarity ($k$-NN) rather than knowledge-unit-based mapping. This demonstrates that: by organizing data into knowledge units, we achieve better alignment with user intent than unstructured semantic retrieval. Additional comparison results for \textit{GPT-5-RAG} and \textit{Embedding-filtering} are provided in Appendix~\ref{sec::supp_variant_baselines}.

\vspace{-0.2cm}

\paragraph{Special-Case Analysis.}
We observed a relatively lower Spearman correlation (0.42) in Material Science compared to other tasks. To investigate this discrepancy, we further evaluated the LLMs using MatSci-NLP~\cite{song2023matsci}, another large-scale expert-curated benchmark in this field (comprising 9,466 classification questions grounded in experimental material science facts).
Interestingly, we found that the alignment between the two expert-curated benchmarks themselves was also limited ($\rho$ = 0.31, $\tau_b$ = 0.22).
This suggests that Material Science is an exceptionally broad and multifaceted discipline~\cite{shoghi2023molecules, wang2025moma}, where different evaluation protocols may focus on orthogonal knowledge areas or distinct reasoning types. We identify the fine-grained unification of such complex and heterogeneous domains as a direction for future research.

\paragraph{Human Evaluation.}
\label{sec::results}
We evaluate the quality of the benchmarks generated by \ourM{} using a random sample of 50 chemistry-related questions that are originally not in multiple-choice format. Three Master students in AI4Chemistry annotate the samples with binary scores for \textit{Correctness} (the validity of the MCQ after transformation) and \textit{Relevance} (the alignment with the benchmark query). The resulting average scores are 0.92 for Correctness and 0.7 for Relevance, confirming the reliability of our benchmark generation pipeline.

\begin{figure*}[t]
\centering	\includegraphics[width=.95\linewidth]{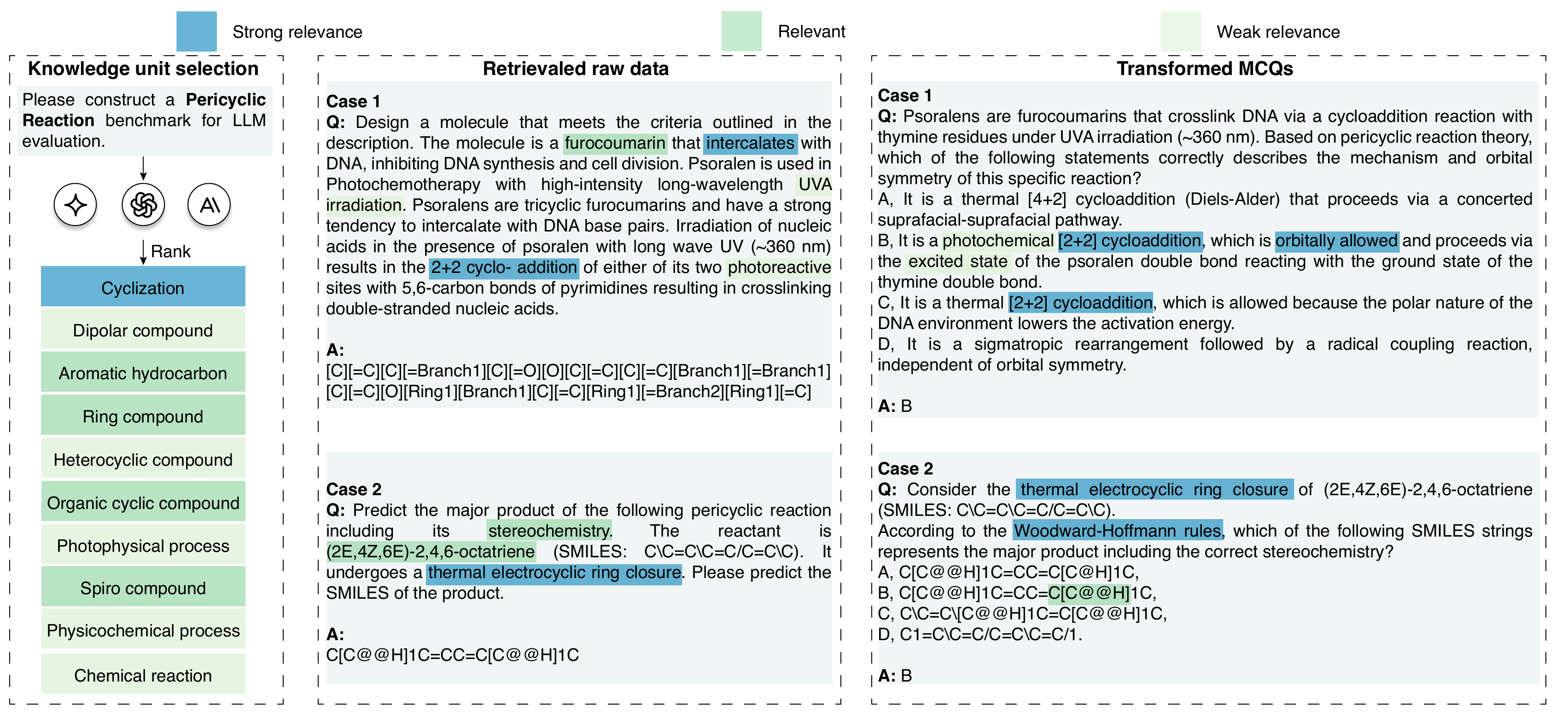}
\vspace{-3mm}
	\caption {Case study on constructing a benchmark for ``Pericyclic Reaction''. The pipeline progresses from identifying relevant knowledge units via voting (Left), retrieving grounded raw data with relevance-highlighted terms (Middle), to generating standardized multiple-choice questions for evaluation (Right)}
\vspace{-4mm}
\label{fig::casestudy}
\end{figure*}

\subsection{Component Analysis}

\paragraph{Effectiveness of Tagger.}
We evaluate the tagger on 1,000 unseen queries generated by compositional sampling over the knowledge units $\mathcal{T}$, where each query is generated based on the keywords corresponding to the knowledge units. We use fuzzy label matching to assess performance, where predicted tags are considered if their normalized Indel similarity with any gold tag exceeds 85\%, using the rapidfuzz library\footnote{\url{https://rapidfuzz.github.io/RapidFuzz/}}. The Macro F1 and Micro F1 scores are 75.2\% and 78.6\%, respectively, demonstrating the effectiveness of the tagger in mapping queries to relevant scientific knowledge.

\paragraph{Effectiveness of Relevance Filtering and Subset Selection Strategies.}
We evaluate the efficacy of our retrieval and selection components by comparing the full \ourM{} framework against two ablation variants: \textit{w/o cutoff} (removing the binary search relevance filter) and \textit{w/o selection} (replacing the intrinsic score-based subset selection with random sampling). The full \ourM{} consistently outperforms both variants across all 11 sub-disciplines (\textbf{Figure~\ref{fig::ablation_slection}}). 
The drop in performance for \textit{w/o selection} (random sampling) is substantial, necessitating the data selection component. Furthermore, removing the binary search cutoff (\textit{w/o cutoff}) also leads to a notable degradation in ranking consistency, suggesting that our cutoff strategy ensures the benchmark remains tightly focused on the target capability. These results demonstrate that both precise relevance filtering and representative subset selection are indispensable for achieving efficient and accurate evaluation.

We further report an automatic evaluation of the quality of LLM-generated distractors in MCQs in Appendix~\ref{sec::supp_distractor_eval}.

\subsection{Benchmark for Pericyclic Reaction}
\label{sec::case_study}
To evaluate the capability of \ourM{} in handling highly specific and novel scientific requirements, we conducted a case study on \textit{Pericyclic Reactions}.
Unlike general organic chemistry queries, pericyclic reactions require a nuanced understanding of orbital symmetry, stereochemistry, and specific reaction conditions (e.g., thermal vs. photochemical)~\cite{WoodwardHoffmann1970}.
No existing LLM benchmark specifically isolates this distinct class of reactions.

As illustrated in \textbf{Figure \ref{fig::casestudy}}, the construction process progresses through three key stages, demonstrating high alignment with expert judgment at each step: Upon receiving the abstract requirement "Pericyclic Reaction benchmark," \ourM{} first decomposes it into fine-grained knowledge anchors.
As shown in the left panel, \ourM{} successfully prioritizes highly relevant concepts.
The ranking produced by \ourM{} highly correlates with expert assessment, where the top-ranked units, such as \textit{Cyclization}, \textit{Aromatic hydrocarbon}, and \textit{Ring compound}, are identified by experts as the most scientifically relevant (indicated by darker shades).
Guided by these concepts, \ourM{} proceeds to data retrieval and question transformation (\textbf{Figure \ref{fig::casestudy}}, middle and right panels). Both the retrieved raw data and the transformed multiple-choice questions exhibit high alignment with expert requirements, accurately capturing critical domain-specific terminology (highlighted in the figure) and complex reaction mechanisms.

\vspace{-1mm}
\section{Related Works}
\vspace{-1mm}
Existing scientific evaluation benchmarks primarily fall into two categories. General scientific benchmarks \cite{MMLU,mmlu-pro,gpqa,scieval,scibench} focus predominantly on evaluating broad reasoning capabilities or generalized scientific common sense, lacking the depth required for specific scientific sub-domains. Conversely, domain-specific benchmarks \cite{chembench,medqa,multimedqa,song2023matsci} offer more specialized evaluations but are inherently static and labor-intensive to maintain due to their reliance on manual expert curation. 

To address the limitations of manual curation, there are several automated benchmarking frameworks in computer vision~\cite{taskmeanything} and general natural language processing~\cite{zeroshot_benchmark,autobencher,sdbench}. They are ill-suited for complex scientific evaluation, as they typically suffer from restricted requirement scopes~\cite{taskmeanything}, a lack of grounded data~\cite{zeroshot_benchmark}, or the inability to reuse compositional knowledge units~\cite{sdbench,autobencher}, thereby failing to meet the critical demands of scientific applications for diversity, intersectionality, and verifiability.

\vspace{-1mm}
\section{Conclusion}
\vspace{-1mm}
In this paper, we introduced \ourM{}, an ontology-driven framework that automates the construction of customized benchmarks for scientific LLM evaluation. By reorganizing scattered scientific corpora into fine-grained, reusable knowledge units, our approach overcomes the scalability bottlenecks of manual curation while ensuring evaluations remain grounded in verifiable facts. Extensive experiments demonstrate that \ourM{} not only achieves high alignment with expert-curated benchmarks in chemistry and healthcare but also reliably assesses model capabilities in novel, unbenchmarked scientific frontiers. This work establishes a scalable and application-aware evaluation framework for navigating the complex and evolving landscape of scientific research.
\section*{Limitations}
\ourM{} has two primary limitations. First, the scope of our current evaluation is constrained by the underlying ontologies, which predominantly cover biomedical and chemical domains (derived from OBO, BioPortal, and OLS). Consequently, some scientific disciplines such as mathematics and theoretical physics, are not yet included. Future work can address this by integrating broader scientific taxonomies to expand the knowledge space.
Second, \ourM{}-curated benchmarks depend on the coverage of the source scientific corpus $\mathcal{D}$. Some knowledge units may currently suffer from data sparsity. With the tagging model, we can identify these low-resource knowledge units by monitoring frequencies of knowledge units. In future work, this framework will continuously evolve and scale as new scientific datasets become available.
\section*{Ethics Statement}
Our experimental evaluation includes a benchmark subset focused on healthcare. We emphasize that our framework and the constructed benchmarks are strictly designed for educational and evaluative purposes within a textual Question-Answering (QA) context. The benchmarks assess an LLM's ability to recall and reason about established scientific facts found in public academic literature. They do not involve, facilitate, or encourage any actionable bio-security threats. Furthermore, all scientific data used in \ourM{} are aggregated from open-source, publicly available repositories. No private, classified, or patient-identifiable data were used in the construction of our benchmarks. We believe that rigorous evaluation of LLMs in these domains is a prerequisite for their safe and responsible deployment in scientific research.
\section*{Acknowledgments}
The authors Ming Zhang, Yiyang Gu, and Junwei Yang are supported by grants from the National Natural Science Foundation of China (NSFC Grant Number 62276002).
The authors Zequn Liu, Yingce Xia, and Shufang Xie are supported by Zhongguancun Academy (Grant No. C20250513 and Grant No. P190260302).
\bibliography{main}
\clearpage
\appendix
\section{Details of Proxy Subset Selection}
\label{app:proxy}
For each data sample $\langle q, a \rangle \in \mathcal{D}_r$, we compute two commonly used intrinsic metrics to characterize its difficulty and quality: (\romannumeral1) \textbf{Hardness Score ($H$):} We utilize open-weights LLMs to calculate the perplexity (PPL) of the answer $a_d$ conditioned on the question $q_d$:
\begin{equation}
    H(d) = \exp\left( -\frac{1}{|a_d|} \sum_{i=1}^{|a_d|} \log P(t_i \mid q_d, t_{<i}) \right)
\end{equation}
where $t_i$ represents the $i$-th token of the answer $a_d$.
(\romannumeral2) \textbf{Quality Score ($Q$):} We assess the linguistic quality using the Flesch Readability Score.

Our objective is to select $\mathcal{P}_q$ such that the distributions of these scores closely approximate those of $\mathcal{D}_q$. We formulate this as minimizing the cumulative Wasserstein distance ($\mathcal{W}$) between the distributions of the subset and the full set:
\begin{equation}
\label{eq::subset_metrics}
    \min_{\mathcal{P}_q} \left( \mathcal{W}(P^{\mathcal{D}_q}_H, P^{\mathcal{P}_q}_H) + \mathcal{W}(P^{\mathcal{D}_q}_Q, P^{\mathcal{P}_q}_Q) \right)
\end{equation}
To solve the sampling of $\mathcal{P}_q$ efficiently, we adopt a cluster-based strategy following SubLIME~\cite{saranathan2025sublime}.
Intuitively, we encode all pairs $d = \langle q, a \rangle$ into an embedding space using RoBERTa~\cite{liu2019roberta}.
We then perform clustering in the embedding space and sample representatives from each cluster.

\section{Details of Ground-Truth Benchmarks}
\label{sec::supp_gt_benchmark}
\textbf{Table~\ref{tab:groundtruth_stats}} presents the detailed statistics of the expert-curated ground-truth benchmarks used in our evaluation.

\begin{table}[h]
    \centering
    \caption{Statistics of the ground-truth benchmarks used in our experiments.}
    \small
    \begin{tabular}{lr}
        \toprule
        \textbf{Task} & \textbf{Size (\# Samples)} \\
        \midrule
        \multicolumn{2}{l}{\textit{\textbf{Chemistry Domain}}} \\
        Analytical Chemistry & 152 \\
        Inorganic Chemistry & 92 \\
        Material Science & 84 \\
        Organic Chemistry & 429 \\
        Physical Chemistry & 165 \\
        Technical Chemistry & 40 \\
        \midrule
        \multicolumn{2}{l}{\textit{\textbf{Healthcare Domain}}} \\
        Virology & 46 \\
        Human Aging & 86 \\
        Medical Genetics & 54 \\
        Anatomy & 79 \\
        Nutrition & 179 \\
        \bottomrule
    \end{tabular}
    \label{tab:groundtruth_stats}
\end{table}
\section{Implementation Details}
\label{sec::supp_implementation}
We finetuned LLaMa-3-8B \cite{llama3} to train a tagging model. We sampled $50,000$ synthetic scientific queries and $30,000$ real-world scientific queries as the training set. The LLM used to construct the dataset is GPT-4o \cite{gpt4}. The model was finetuned for $2$ epochs with a learning rate of $2e-5$. We conducted experiments on 8 NVIDIA A100 GPUs in a standard Linux environment, using the vLLM~\cite{kwon2023efficient} framework for efficient large-scale model inference.

We constructed a comprehensive scientific data corpus by aggregating diverse high-quality instruction-tuning datasets and benchmarks, including \textit{SciRIFF} \cite{wadden2025sciriff}, \textit{SciInstruct} \cite{zhang2024sciinstruct}, \textit{Mol-Instruct} \cite{mol-instructions}, \textit{MultiMedQA} \cite{multimedqa}, \textit{SciEval} \cite{scieval}, \textit{MMLU-Pro}~\cite{mmlu-pro}, \textit{GPQA}~\cite{gpqa}, \textit{IfBench}~\cite{ifbench}, and \textit{SimpleQA}~\cite{simpleqa}. This collection yielded a total of $2,000,367$ data instances. To avoid data leakage, we filtered out any instances that overlap with the expert-curated ground-truth benchmarks used in our experiments.

To facilitate multi-model voting as discussed in both Section \ref{sec:voting} and Section \ref{sec:benchmark generation}, we utilized three advanced models: GPT-5, Claude-opus-4.5, and Gemini-3-pro-preview.
We set the voting parameter $K_1 = 10$.
For the proxy subset selection, we set $K_2=100$, ensuring the scale remains consistent with the expert-curated ground-truth benchmarks (Appendix \ref{sec::supp_gt_benchmark}).
Furthermore, following the SubLIME~\cite{saranathan2025sublime} framework, we employed a cluster-based strategy to sample $100$ candidate sets and selected the optimal subset based on the criteria defined in Equation \ref{eq::subset_metrics}.

We evaluated 10 LLMs, including Gemini-2.5-flash,
  GPT-4o,
  GPT-5-chat,
  Claude-opus-4.5,
  Qwen3-235b-a22b-instruct-2507,
  Qwen3-max,
  Grok-4-fast-non-reasoning,
  Kimi-k2-preview,
  DeepSeek-V3.2 and
  Mistral-large on the all comparison benchmarks.

\begin{table*}[!t]
\centering
\small
\caption{Comparison between a greedy retrieval strategy and \ourM{} on healthcare benchmarks. Scores are Spearman correlations with the corresponding expert-curated ground-truth rankings.}
\resizebox{0.65\linewidth}{!}{
\begin{tabular}{c|ccccc}
\toprule
Method & Virology & Human aging & Medical genetics & Anatomy & Nutrition \\
\midrule
Greedy Search & 0.21 & 0.34 & 0.16 & 0.24 & 0.31 \\
\ourM{} & \textbf{0.55} & \textbf{0.49} & \textbf{0.42} & \textbf{0.62} & \textbf{0.78} \\
\bottomrule
\end{tabular}
}
\label{tab:supp_greedy_search}
\end{table*}

\begin{table*}[!t]
\centering
\small
\caption{Ranking consistency analysis of 10 LLMs on 6 chemistry-specific benchmarks. The table compares \ourM{} against baseline benchmarks in preserving model rankings, quantified by Spearman correlation coefficients.
Higher scores indicate better consistency. Best results are in \textbf{bold}.}
\resizebox{\linewidth}{!}{
\begin{tabular}{c|cccccc}
\toprule
Method & Analytical chemistry & Inorganic chemistry & Material science & Organic chemistry & Physical chemistry & Technical chemistry \\
\midrule
GPT-5 & -0.11 & 0.05 & -0.04 & 0.38 & 0.24 & -0.07 \\
GPT-5-RAG & 0.01 & -0.02 & -0.07 & 0.41 & 0.35 & -0.02 \\
Embedding & -0.34 & -0.59 & -0.39 & 0.11 & -0.73 & -0.41 \\
Embedding-filtering & 0.23 & 0.04 & 0.15 & 0.37 & 0.04 & 0.11 \\
\ourM{} & \textbf{0.86} & \textbf{0.67} & \textbf{0.42} & \textbf{0.89} & \textbf{0.74} & \textbf{0.86} \\
\bottomrule
\end{tabular}
}
\label{tab:supp_variant_baselines_chem}
\end{table*}

\begin{table*}[!t]
\centering
\small
\caption{Ranking consistency analysis of 10 LLMs on 5 healthcare-specific benchmarks. The table compares \ourM{} against baseline benchmarks in preserving model rankings, quantified by Spearman correlation coefficients. Higher scores indicate better consistency. Best results are in \textbf{bold}.}
\resizebox{0.65\linewidth}{!}{
\begin{tabular}{c|ccccc}
\toprule
Method & Virology & Human aging & Medical genetics & Anatomy & Nutrition \\
\midrule
GPT-5 & 0.25 & 0.20 & 0.09 & 0.11 & 0.52 \\
GPT-5-RAG & 0.31 & 0.26 & 0.02 & 0.09 & 0.48 \\
Embedding & 0.18 & 0.21 & -0.21 & -0.32 & 0.27 \\
Embedding-filtering & 0.24 & 0.34 & 0.13 & 0.26 & 0.51 \\
\ourM{} & \textbf{0.55} & \textbf{0.49} & \textbf{0.42} & \textbf{0.62} & \textbf{0.78} \\
\bottomrule
\end{tabular}
}
\label{tab:supp_variant_baselines_health}
\end{table*}

\paragraph{Discussion on Non-Monotonicity and Selection Strategy} We acknowledge that the relevance of data samples within the sorted list $L_{\mathcal{D}'_r}$ does not strictly follow a monotonic decreasing trend, which theoretically challenges the standard binary search assumption. A seemingly more rigorous alternative would be a "Greedy Top-$K$ Selection", where we sequentially scan $L_{\mathcal{D}'_r}$ from the beginning and select the first $K_2$ samples judged as relevant by the voting ensemble. 
However, our experiments show that this greedy strategy ($K_2 = 100$) results in a significantly lower correlation with ground-truth benchmarks compared to the proposed binary search method (\textbf{Table~\ref{tab:supp_greedy_search}}). 
We hypothesize that the top-ranked instances in $L_{\mathcal{D}'_r}$, which share the highest semantic and keyword overlap with the requirement, often correspond to "canonical" textbook knowledge or highly frequent scientific facts. These instances are likely exposed during the pre-training of most LLMs (potential data leakage), leading to a performance ceiling effect where most models achieve near-perfect accuracy. Consequently, these "highly relevant" samples lack the discriminative power to differentiate model capabilities. In contrast, our binary search strategy determines a cutoff based on a broader structural trend. This approach inadvertently bypasses the saturation of the top-most samples and captures a more challenging and diverse distribution of data, thereby offering higher discriminative validity and better alignment with expert rankings.

\section{More Experiment Results}
\label{sec::supp_variant_baselines}

To further isolate the contribution of our ontology-driven architecture, we report Spearman correlation results for two alternative baselines. \textit{GPT-5-RAG} augments the synthetic \textit{GPT-5} generator with open-book retrieval from the full corpus, while \textit{Embedding-filtering} applies the same binary-search cutoff and proxy subset selection used in \ourM{} on top of the \textit{Embedding} candidate pool.
As shown in \textbf{Table~\ref{tab:supp_variant_baselines_chem}} and \textbf{Table~\ref{tab:supp_variant_baselines_health}}, adding retrieval to \textit{GPT-5} yields only limited gains over the fully synthetic baseline, while \textit{Embedding-filtering} improves over raw \textit{Embedding} but still remains consistently below \ourM{} across both chemistry and healthcare tasks. These results show that neither retrieval augmentation alone nor post-retrieval filtering alone is sufficient to match the performance of the full ontology-grounded pipeline. This confirms that the fine-grained structural guidance provided by our ontology-grounded knowledge units is indispensable for accurately capturing complex scientific requirements.

\section{Details of Human Evaluation}
\subsection{Human Annotation Guidelines}

\textbf{Objective:} You will be presented with 50 multiple-choice questions generated for a specific scientific requirement (e.g., \textit{Technical Chemistry}). For each question, please provide two binary labels (\texttt{0} or \texttt{1}) based on the criteria below.

\subsubsection*{1. Relevance Label (\texttt{relevant})}
Determine whether the question requires specific knowledge of the target requirement.

\begin{itemize}
    \item \textbf{1 (Relevant):} The question is strictly aligned with the requirement. 
    \item \textbf{0 (Irrelevant):} The question is off-topic, generic (can be answered by a layperson), or belongs to a distinctly different scientific field.
\end{itemize}

\subsubsection*{2. Correctness Label (\texttt{correct})}
Evaluate the scientific accuracy of the MCQs.

\begin{itemize}
    \item \textbf{1 (Correct):} The questions and options are scientifically accurate.
    \item \textbf{0 (Incorrect):} The question is wrong, or the selected option is factually wrong, scientifically flawed, or there is a significantly better/more accurate option available in the choices.
\end{itemize}
\subsection{Human Annotators}
The annotation was performed by three Master's students specializing in AI for Science (AI4Chemistry). Participants were recruited from an university department and were compensated at an hourly rate exceeding the local minimum wage. The results are discussed in Section \ref{sec::results}.

For the case study (Section \ref{sec::case_study}), we engaged a Ph.D. researcher specializing in AI and Chemistry to assess the generated benchmark. The expert was tasked with categorizing the knowledge units and keywords in the MCQs into three distinct levels of granularity and relevance: \textit{Strong Relevance}, \textit{Relevant}, and \textit{Weak Relevance}. The expert was compensated at a professional hourly rate commensurate. 

Prior to the task, all annotators were informed of the data usage scope and provided written informed consent. The study protocol was reviewed and determined exempt by an ethics review board. 

\section{Automatic Evaluation of Distractor Quality}
\label{sec::supp_distractor_eval}
To validate the quality of LLM-generated distractors, we evaluate 500 generated MCQs with Claude-Opus-4.6 as an independent judge. Each MCQ is scored on a 1--10 scale along two dimensions: \textbf{Heuristic Transparency Score (HTS)}, which measures whether distractors can be eliminated by superficial linguistic cues, and \textbf{Scientific Plausibility Score (SPS)}, which measures how scientifically relevant and challenging the distractors are. The generated MCQs obtain an average HTS of 8.7 and SPS of 7.9, indicating low heuristic leakage and strong near-miss quality. We additionally construct a context-free baseline where the judge only sees the options and a simplified question, without the supporting scientific context; accuracy drops to 22\%, close to random guess. These results support that our MCQ transformation pipeline produces distractors that are both scientifically plausible and difficult to solve through shallow heuristics alone.

\section{Prompt Template Inventory}
\label{sec::supp_prompt}

\begin{tcolorbox}[
breakable,
title = Granularity classification of scientific terms,
colback=white,
colframe=gray!75!black,
enhanced
]
\label{prompt::subfield-granularity}
\small
\textbf{\texttt{System:}} You are a helpful AI assistant. \\
\\
\textbf{\texttt{User:}} Determine whether the given term is a suitable scientific subfield and answer with one of the following categories: \\
\\
\textbf{(moderate)}: The term is appropriately specific for a scientific subfield. It refers to a category that is neither too general nor too specific for its scientific context. Its scale is similar to the scale of chapter names in subject textbooks. \\
\textbf{(too coarse)}: The term is overly broad or vague for a scientific subfield. It encompasses a wide range of concepts that could be divided into smaller, more specific subfields. \\
\textbf{(too fine)}: The term is overly specific and pertains to a very narrow aspect of a scientific subfield. It may be too detailed to serve as a broader category within the discipline. \\
\\
Answer at the beginning, explain later. The examples are as follows:\\
\\
Example 1: \\
\textit{Input: term: anatomical entity} \\
\textit{Output: (moderate); Explanation: Anatomical entity refers to a category that is sufficiently specific for many biological subfields but not too narrow.} \\
\\
Example 2: \\
\textit{Input: term: nuclear structure} \\
\textit{Output: (moderate); Explanation: Nuclear structure is a well-defined category in molecular biology, specific but not too narrow.} \\
\\
Example 3: \\
\textit{Input: term: electronic file status} \\
\textit{Output: (too fine); Explanation: The term refers to a very specific technical concept that is too detailed to be considered a scientific subfield.} \\
\\
Example 4: \\
\textit{Input: term: b-lymphocyte} \\
\textit{Output: (too fine); Explanation: While important, the term refers to a very specific type of cell, not a broad enough category to encompass a subfield.} \\
\\
Example 5: \\
\textit{Input: term: continuant} \\
\textit{Output: (too coarse); Explanation: The term is too general and could refer to a wide variety of objects or concepts, making it too broad for a specific subfield.} \\
\\
Example 6: \\
\textit{Input: term: occurrent} \\
\textit{Output: (too coarse); Explanation: Occurrent is overly vague and applies to many concepts, making it too broad for a scientific subfield.} \\
\\
\textit{Input: term: \{term\}.}
\end{tcolorbox}

\begin{tcolorbox}[
breakable,
title = Query generation,
colback=white,
colframe=gray!75!black,
enhanced
]
\small
\textbf{\texttt{System:}} You are a helpful AI assistant. \\

\textbf{\texttt{User:}}\\
\textbf{Low Complexity:} You are a \{persona\}. Generate a user query containing the following keywords: \{keywords\}. Do not introduce other scientific entities or topics. Only return the query. \\
\textbf{High Complexity:} You are a \{persona\}. Generate a user query containing the following keywords: \{keywords\}. Do not introduce other scientific entities or topics. Make the query long and complex. Only return the query. \\

\textit{Scientific Personas:} \\
- Astrophysicist \\
- Marine Biologist \\
- AI Researcher \\
- Molecular Geneticist \\
- Quantum Physicist \\
- Environmental Chemist \\
- Neuroscientist \\
- Ecologist \\
- Bioinformatician \\
- Pharmacologist \\
- Geologist \\
- Biomedical Engineer \\
- Mathematical Modeler \\
- Virologist \\
- Behavioral Psychologist \\
- Data Scientist \\
- Theoretical Chemist \\
- Climate Scientist \\
- Structural Biologist \\
- Robotics Engineer

\end{tcolorbox}

\begin{tcolorbox}[
breakable,
title = Benchmark requirements and descriptions,
colback=white,
colframe=gray!75!black,
enhanced
]
\small
\textbf{Chemistry benchmarks:}
\begin{itemize}[leftmargin=*]
    \item \textit{Analytical chemistry}: Generate questions that test knowledge and reasoning in analytical chemistry.
The questions should assess understanding of how experimental analytical signals 
(e.g., NMR, IR, UV–Vis, mass spectra, chromatographic behavior, titration curves)
relate to molecular structure, composition, concentration, or purity.
Focus on conceptual interpretation and chemical reasoning rather than numerical 
data processing or instrument-specific operating procedures.
    \item \textit{Inorganic chemistry}: Generate questions that test core knowledge and reasoning in inorganic chemistry.
The questions should focus on electronic structure, oxidation states, coordination 
geometry, ligand field effects, symmetry, and periodic trends in inorganic systems.
Emphasize conceptual understanding of structure–property relationships rather than 
memorization of isolated facts.
    \item \textit{Material science}: Generate questions that evaluate understanding in materials science.
The questions should assess how atomic or microstructural features 
(e.g., crystal structure, defects, phases, interfaces) determine macroscopic 
properties such as mechanical strength, electrical conductivity, or thermal behavior.
Focus on structure–property reasoning rather than detailed synthesis protocols.
    \item \textit{Organic chemistry}: The organic chemistry benchmark assesses a wide range of skills on reasoning about chemical structures and reaction pathways, such as Reaction Mechanism Identification, Product Prediction, NMR Signal Prediction, Number of Isomers, Polymer Chemistry, Nomenclature Conversion and Organic Reactivity.
    \item \textit{Physical chemistry}: Generate questions that test conceptual understanding in physical chemistry.
The questions should assess reasoning about thermodynamics, kinetics, 
equilibrium, and molecular-level physical principles.
Emphasize qualitative reasoning about trends and relationships rather than 
explicit numerical calculation.
    \item \textit{Technical chemistry}: Generate questions that assess knowledge in technical and industrial chemistry.
The questions should focus on chemical processes at scale, such as reactor behavior,
process optimization, safety considerations, and material or energy efficiency.
Emphasize reasoning about system-level behavior rather than detailed engineering design.
\end{itemize}

\textbf{Healthcare benchmarks:}
\begin{itemize}[leftmargin=*]
    \item \textit{Virology}: Generate questions that test conceptual understanding in virology.
The questions should assess knowledge of viral structure, replication cycles,
genome organization, and interactions with host cells and immune systems.
Avoid clinical treatment guidelines or laboratory diagnostic protocols.
    \item \textit{Human aging}: Generate questions that probe understanding of biological mechanisms of human aging.
The questions should focus on molecular, cellular, and systemic processes 
associated with aging, such as genomic stability, cellular senescence, 
metabolic regulation, and tissue-level decline.
Emphasize mechanistic reasoning rather than epidemiological statistics.
    \item \textit{Medical genetics}: Generate questions that test reasoning in medical genetics.
The questions should assess understanding of inheritance patterns, 
genotype–phenotype relationships, penetrance, and genetic variation.
Focus on conceptual genetic reasoning rather than clinical decision-making.
    \item \textit{Anatomy}: Generate questions that evaluate knowledge of human anatomy.
The questions should focus on the identification, spatial relationships, 
and functional roles of anatomical structures.
Avoid surgical procedures or pathological conditions.
    \item \textit{Nutrition}: Generate questions that assess understanding of nutritional science.
The questions should focus on the biological roles of macro- and micronutrients,
their involvement in metabolism, and the physiological consequences of deficiency 
or imbalance.
Emphasize mechanistic understanding over dietary recommendations.
\end{itemize}
\end{tcolorbox}

\begin{tcolorbox}[
breakable,
title = Voting-based relevant tag selection,
colback=white,
colframe=gray!75!black,
enhanced
]
\small
\textbf{\texttt{System:}} You are an expert in \textbf{\{domain\}}. Your task is to map a benchmark description to the most relevant technical tags \\

\textbf{\texttt{User:}}
Task:\\
Given a \textbf{\{domain\}} benchmark description, identify and rank the most relevant tags from a candidate list.\\

Benchmark Description: \textbf{\{description\}} \\

Candidate Tags (sorted by frequency, lowest frequency first;
lower frequency usually indicates higher specificity): \textbf{\{Tag list\}}. \\

\textbf{Ranking Principles:} Rank tags from highest to lowest relevance to the benchmark, following these rules: \\

Relevance First: \\
A tag is relevant if it directly reflects the core concepts, tasks, data modalities, or evaluation focus of the benchmark. Irrelevant or weakly related tags should not be selected. \\

Specificity as a Tie-breaker: \\
If multiple tags are similarly relevant, rank the more specific and narrowly scoped tag higher. Prefer concrete technical terms (e.g., “Histone Acetylation Prediction”) over broader categories (e.g., “Epigenetics”). \\

Avoid Overly Generic Tags:\\
High-level or generic tags (e.g., “biological process”, “chemical entity”) should only be selected if no more specific alternative applies. \\

Frequency Awareness: \\
When relevance and specificity are comparable, prefer lower-frequency tags, as they tend to be more precise. \\

Output Requirements: \\ 
Return a single list of tags, sorted from most to least relevant. For efficiency, return only the top 100 tags (or fewer if fewer are relevant). Do not include explanations, scores, or extra text—output the ranked list only.
\end{tcolorbox}

\begin{tcolorbox}[
breakable,
title = Benchmark generation,
colback=white,
colframe=gray!75!black,
enhanced
]
\label{prompt::gpt-5-benchmark-generation}
\small
\textbf{\texttt{System:}} You are an expert in \textbf{\{domain}\} and tasked with constructing a high-quality benchmark to assess the domain-specific knowledge abilities of large language models. Please return the benchmark in a JSON format.  \\
\\
\textbf{\texttt{User:}} Your task is to generate exactly \textbf{\{K\}} single-choice questions in the domain of \textbf{\{domain\}}.  \\
Detailed description of this domain: \textbf{\{description\}} \\
The questions should: \\
1. Focus on core concepts, expert-level knowledge, and non-trivial reasoning in this domain. \\
2. Avoid trivial definitions, purely factual memorization, or overly ambiguous questions. \\
3. Include a mix of: \\
- Conceptual understanding \\ 
- Mechanism or principle-based reasoning \\
- Application or scenario-based reasoning \\
4. Be answerable without external tools, but not solvable by surface-level pattern matching. \\
\\
Question format: \\
1. Each question must have 4–5 options. \\
2. Options should be concise and mutually exclusive. \\
3. Each question have only one correct answers. \\
\\
Output format (STRICT): \\
Return only a JSON array of length \textbf{\{K\}}. \\
Each element must have the following structure:\\
\{\{\\
"query": "<question text with options labeled A, B, C, D (and E if applicable)>", \\
"answer": "<correct option label>" \\
\}\}
\end{tcolorbox}

\begin{tcolorbox}[
breakable,
title = MCQ transformation,
colback=white,
colframe=gray!75!black,
enhanced
]
\small
\textbf{\texttt{System:}} You are an expert in \textbf{\{domain\}} and tasked with curating a rigorous benchmark to evaluate the capabilities of Large Language Models. Please return the processed entry in a JSON format. \\
\\
\textbf{\texttt{User:}} Your task is to convert the following raw problem content into a standardized single-choice question suitable for LLM evaluation. \\
Raw problem: \textbf{\{input\_content\}} \\
\\
Conversion Guidelines: \\
1. Format Adaptation: \\
- If the input is already a multiple-choice question: Preserve the original stem and options exactly. Ensure the formatting aligns with the output requirements. \\
- If the input is not a multiple-choice question: Convert it into a single-choice question by generating 3–4 incorrect options (distractors). \\
2. Distractor Engineering: \\
- Avoid trivial errors, logical fallacies that are easily filtered, or clearly unrelated concepts. \\
3. Fidelity \& Difficulty: \\
- Strict adherence to the factual truth and reasoning logic of the original content is required. \\
- Do not simplify the problem complexity. The resulting MCQ must maintain the same discriminative power as the original input. \\
4. Exclusivity: Ensure there is exactly one indisputably correct option. \\
\\
Question format: \\
1. The final output must contain 4–5 options (A, B, C, D, [E]). \\
2. Options should be concise and mutually exclusive. \\
\\
Output format (STRICT): \\
Return only a single JSON object. \\
The object must have the following structure: \\
\{\{\\
"query": "<question stem followed by options labeled A, B, C, D (and E if applicable), separated by newlines>", \\
"answer": "<correct option label, e.g., 'A'>" \\
\}\}
\end{tcolorbox}

\section{LLMs Usage}

We adhere to the ACL Code of Ethics. We use large language models solely for polishing writing. All scientific contributions remain entirely our own.

\end{document}